\documentclass[10pt,twocolumn,letterpaper]{article}

\usepackage[cvprfinal]{cvpr}

\usepackage{graphicx}
\usepackage{amsmath}
\usepackage{amssymb}
\usepackage{booktabs}
\usepackage[usenames,dvipsnames]{xcolor}
\usepackage{url}
\usepackage{enumitem}
\usepackage{xspace}
\usepackage{paralist}
\usepackage{multirow}
\usepackage[numbers,sort&compress]{natbib}

\usepackage[pagebackref,breaklinks,colorlinks]{hyperref}

\makeatletter

\newcommand{\Rmnum}[1]{\expandafter\@slowromancap\romannumeral #1@}
\makeatother

\usepackage{bm}

\newif\ifshowcomments
\showcommentstrue %

\ifshowcomments
    \newcommand{\todo}[1]{\noindent\textcolor{BrickRed}{\textbf{[TODO:~}#1\textbf{]}}}
    \newcommand{\otmar}[1]{\noindent\textcolor{ForestGreen}{\textbf{[Otmar:~}#1\textbf{]}}}
    
    \newcommand{\jie}[1]{\noindent\textcolor{Dandelion}{\textbf{[Jie:~}#1\textbf{]}}}
    \newcommand{\MK}[1]{\noindent\textcolor{Dandelion}{\textbf{[Manuel:~}#1\textbf{]}}}
    \newcommand{\OH}[1]{{\color{blue}[OH: #1]}}
    \newcommand{\pmnote}[1]{\PM{#1}}
    \newcommand{\oh}[1]{\OH{#1}}
    \newcommand{\cg}[1]{{\color{magenta}[cg: #1]}}
    \newcommand{\JZ}[1]{{\color{Purple}[Juan: #1]}}
    
\else
    \newcommand{\todo}[1]{\unskip}
    \newcommand{\otmar}[1]{\unskip}
    \newcommand{\xu}[1]{\unskip}
    \newcommand{\jie}[1]{\unskip}
    \newcommand{\MK}[1]{\unskip}
    \newcommand{\OH}[1]{\unskip}
    \newcommand{\pmnote}[1]{\unskip}
    \newcommand{\oh}[1]{\unskip}
    \newcommand{\cg}[1]{\unskip}
    \newcommand{\JZ}[1]{\unskip}

\fi    
    
\newcommand{\methodname}{X-Avatar\xspace}
\newcommand{\methodnames}{X-Avatars\xspace}
\newcommand{\suppmat}{Supp. Mat\xspace}
\newcommand{\ourdataset}{X-Humans\xspace}

\newcommand{\figref}[1]{Fig.~\ref{#1}}
\newcommand{\tabref}[1]{Tab.~\ref{#1}}
\newcommand{\secref}[1]{Sec.~\ref{#1}}

\DeclareMathOperator*{\argmax}{arg\,max}

\newcommand{\pose}{\boldsymbol{\theta}}
\newcommand{\shape}{\boldsymbol{\beta}}
\newcommand{\face}{\boldsymbol{\psi}}

\definecolor{babyblue}{rgb}{0.54, 0.81, 0.94}

\usepackage[capitalize]{cleveref}
\crefname{section}{Sec.}{Secs.}
\Crefname{section}{Section}{Sections}
\Crefname{table}{Table}{Tables}
\crefname{table}{Tab.}{Tabs.}

\def\cvprPaperID{} 
\def\confName{CVPR}
\def\confYear{2023}

\begin{document}

\title{\methodname: Expressive Human Avatars}
\vspace{-1em}
\author{Kaiyue Shen$^{*1}$ \quad  Chen Guo$^{*1}$ \quad Manuel Kaufmann$^{1}$ \quad Juan Jose Zarate$^{1}$ \\ Julien Valentin$^{2}$ \quad Jie Song $^{1}$ \quad Otmar Hilliges $^{1}$ \\$^1$ETH Z{\"u}rich \quad $^2$Microsoft
}

\twocolumn[{%
\renewcommand\twocolumn[1][]{#1}%
\maketitle

\vspace{-3.5em}
\begin{center}
    \captionsetup{type=figure}
   \includegraphics[width=\linewidth,trim=6 0 6 10,clip]{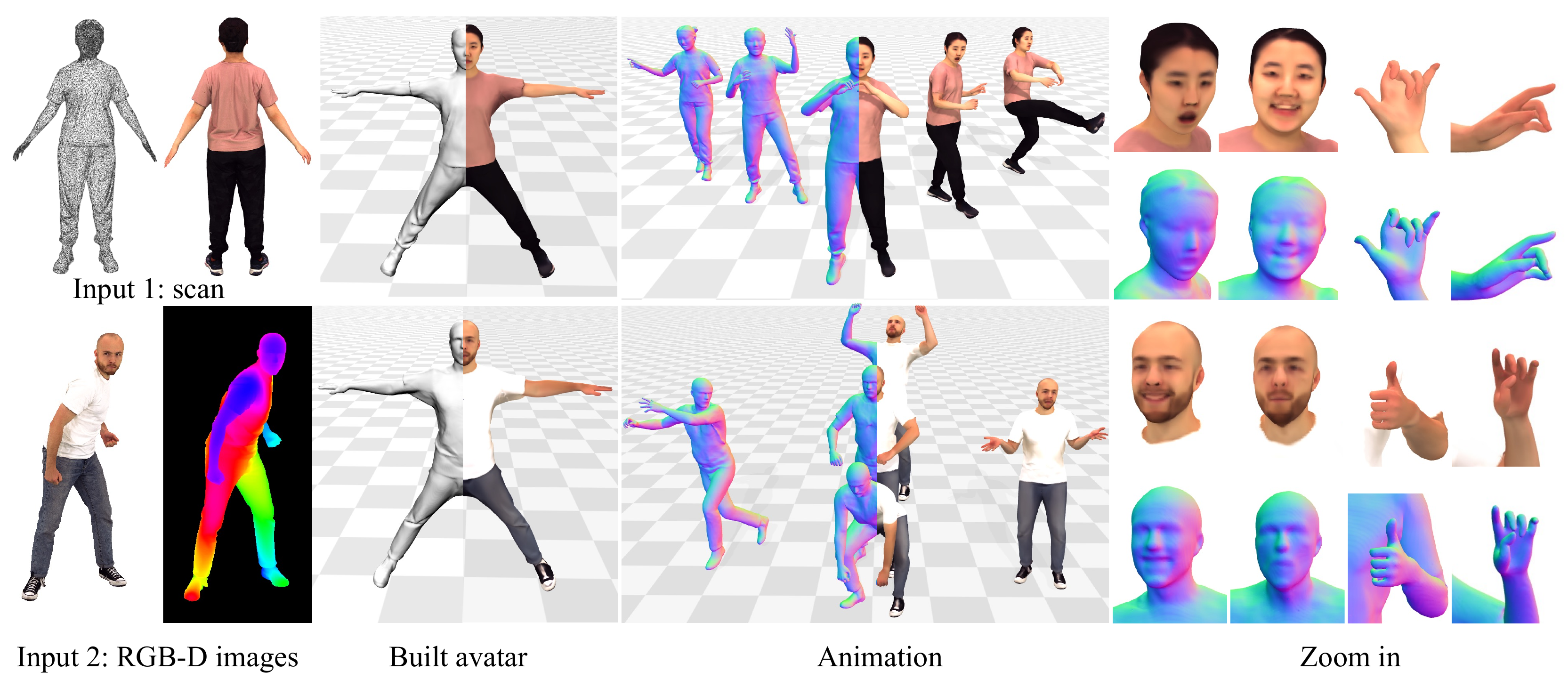}
    \captionof{figure}{We propose \textbf{\methodname}, an animatible implicit human avatar model capable of capturing human body pose, hand pose, facial expressions, and appearance. \methodname can be created from input 3D scans (top row) or RGB-D images (bottom row) and displays high-quality geometry as well as appearance under animation. \methodname captures facial expressions and hand gestures (right), making it the first implicit human avatar model to capture the richness of the human state in a unified model.
    }\label{fig:teaser}
    
\end{center}%

}]

\def\thefootnote{*}\footnotetext{These authors contributed equally to this work}
\newcommand{\figurePipeline}{

\begin{figure*}[t]

\includegraphics[width=\linewidth]{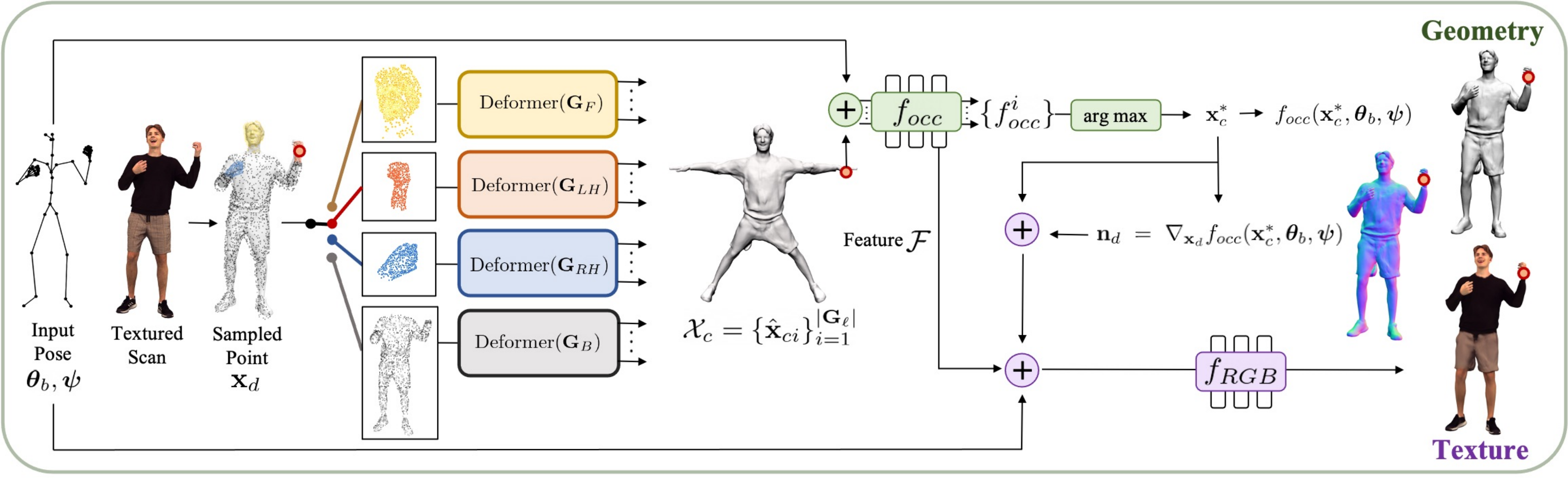}

\caption{\textbf{Method Overview.} Given a posed scan with an SMPL-X registration, we first adaptively sample points $\mathbf{x}_d$ in deformed space per body part $\ell$ (face $F$, left hand $LH$, right hand $RH$, body $B$). A part-specific deformer network finds the corresponding candidate points $\hat{\mathbf{x}}_{ci}$ (for $1 \leqslant i \leqslant |\mathbf{G}_\ell|)$ in canonical space via iterative root finding. The deformers share the parameters of the skinning network, but each deformer is initialized with only the bone transformations $\mathbf{G}_{\ell}$ (cf. \figref{fig:deformer}). The final shape is obtained via an occupancy network $f_{occ}$. We further model appearance via a texture network that takes as input the body pose $\pose_b$, facial expression $\face$, the last layer $\mathcal{F}$ of $f_{occ}$, the canonical point $\mathbf{x}^{*}_c$ and the normals $\mathbf{n}_d$ in deformed space. The normals correspond to the gradient $\nabla_{\mathbf{x}_d}f_{occ}(\mathbf{x}_c^{*}, \pose_b, \face)$.
}
\label{fig:pipeline}
\end{figure*}
}

\newcommand{\figureDeformer}{

\begin{figure}
    \centering
    \includegraphics[width=\linewidth]{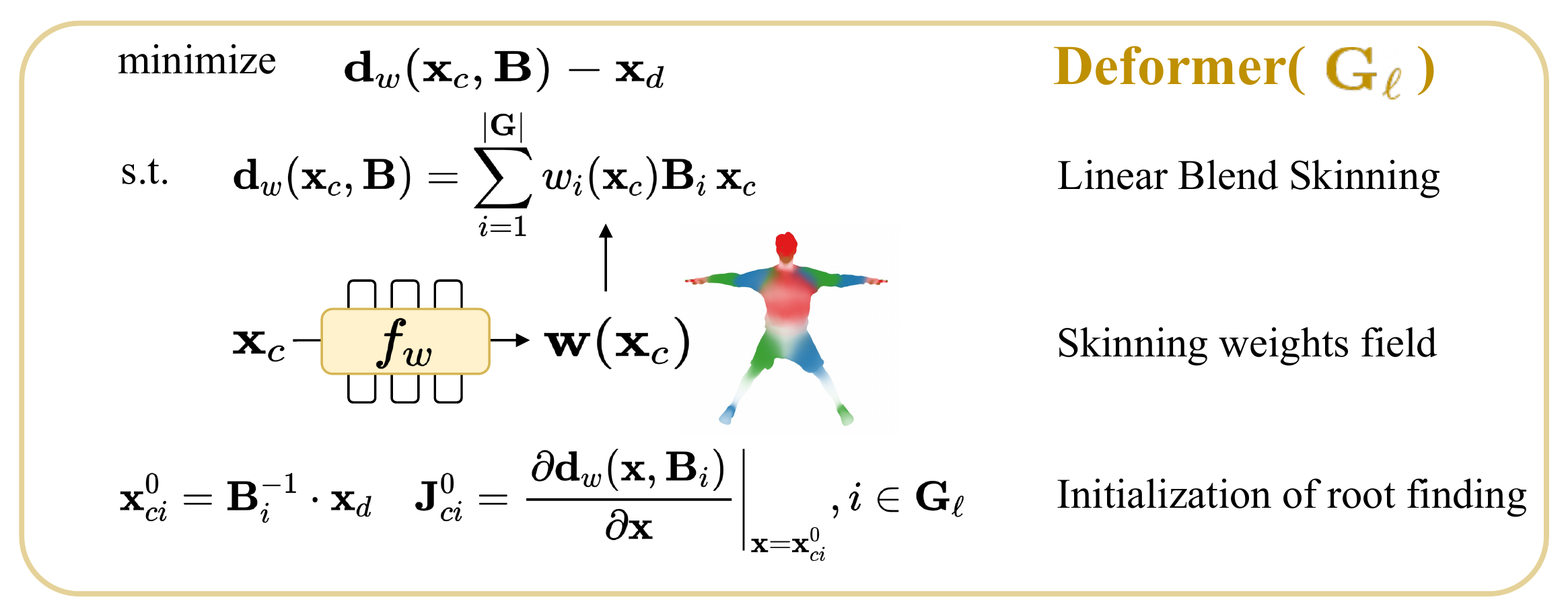}
    \caption{\textbf{Part-specific Deformer.} Each deformer shown in \figref{fig:pipeline} is initialized with the bone transformations belonging to a specific part $\mathbf{G}_{\ell}, \ell \in \{F, LH, RH, B\}$, but shares the parameters of $f_w$.}
    \label{fig:deformer}
\end{figure}

}

\newcommand{\figureablationhandgeometry}{

\begin{figure*}[t]
\raggedleft
\includegraphics[width=\linewidth,trim=0 3 0 0,clip]{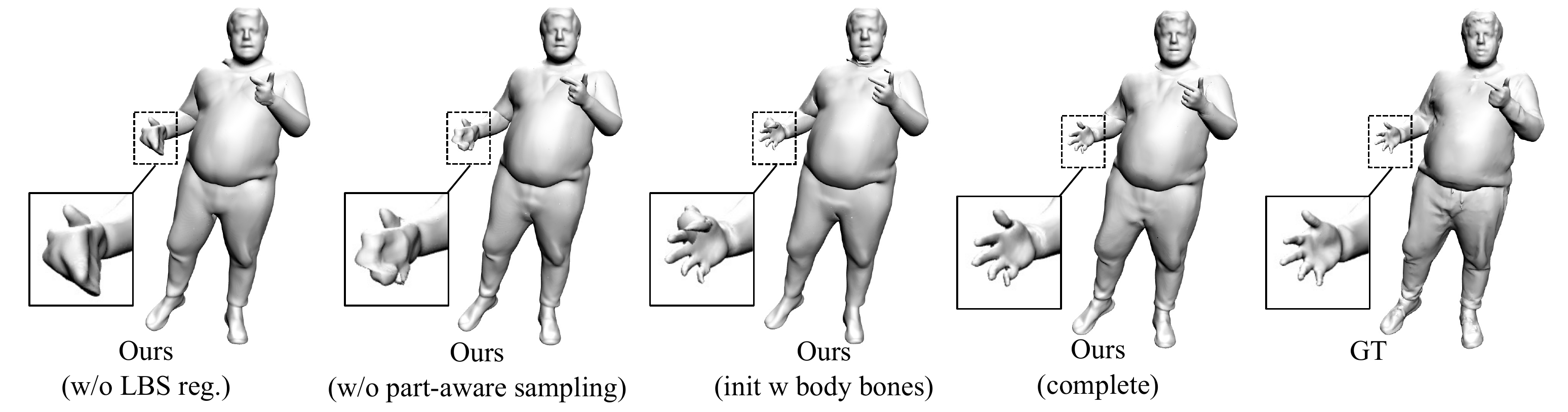}

\vspace{-1em}
\caption{\textbf{Effect of our design decisions on the resulting geometry.} Notice how all baselines struggle to recover accurate hand geometry.}

\label{fig:ablation_hand}
\end{figure*}
} 

\newcommand{\figureablationpartawaresampling}{

\begin{figure}[t]
\raggedleft
\includegraphics[width=\linewidth,trim=0 3 0 0,clip]{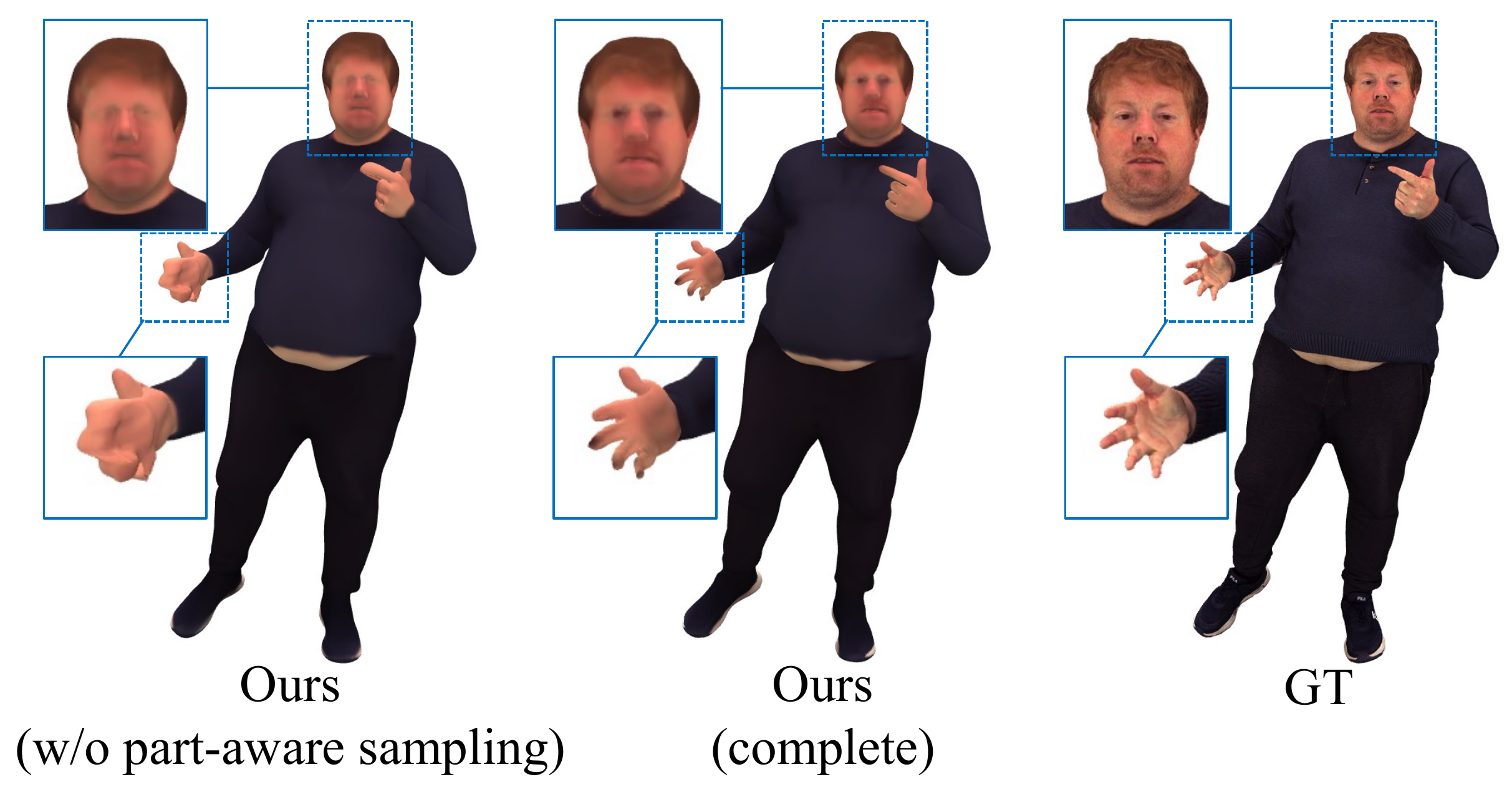}

\vspace{-1em}
\caption{\textbf{Effect of our part-aware sampling strategy} on the hand geometry and texture prediction of the face.}

\label{fig:ablation_part_aware_sampling}
\end{figure}
}

\newcommand{\figurecmpGRAB}{

\begin{figure}[t]
\raggedleft
\includegraphics[width=\linewidth,trim=0 0 0 0,clip]{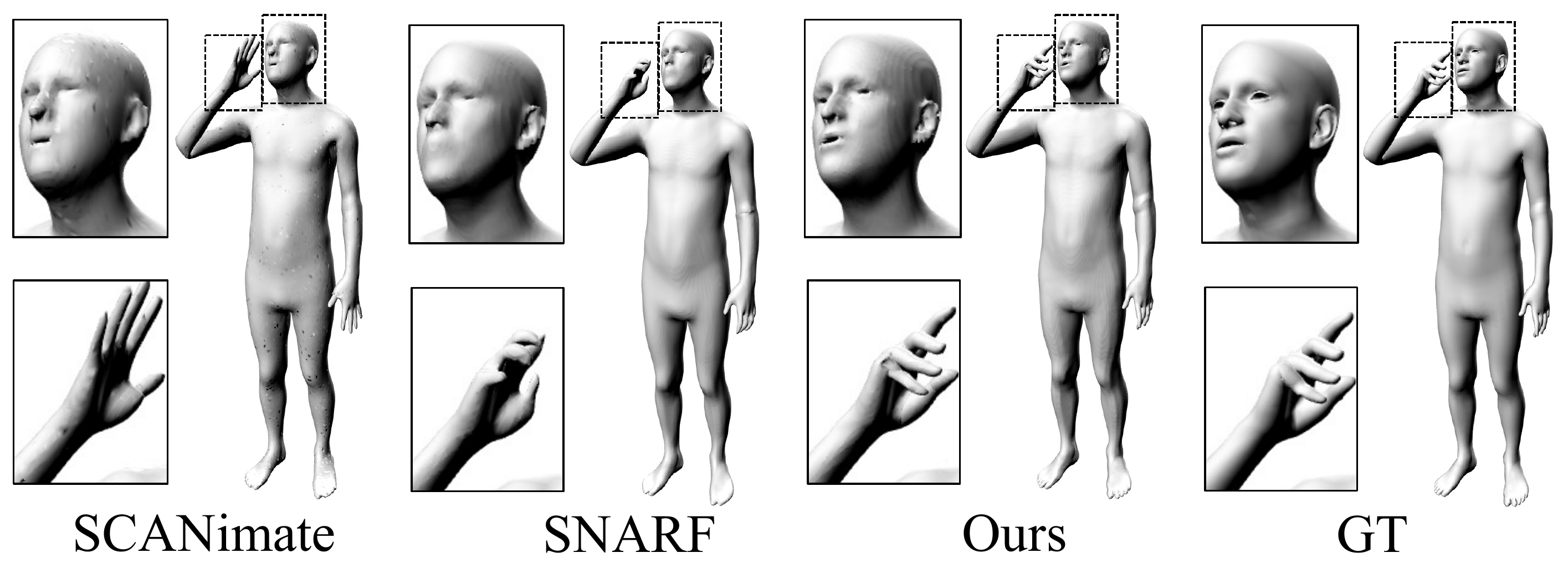}
\vspace{-2em}

\caption{\textbf{Qualitative results on GRAB dataset}. Our method recovers hand articulation and facial expression most accurately.}
\label{fig:cmp_GRAB}
\end{figure}
}

\newcommand{\figurecmpRealscan}{

\begin{figure*}[t]
\begin{center}
\raggedleft
\includegraphics[width=\linewidth,trim=0 0 0 0,clip]{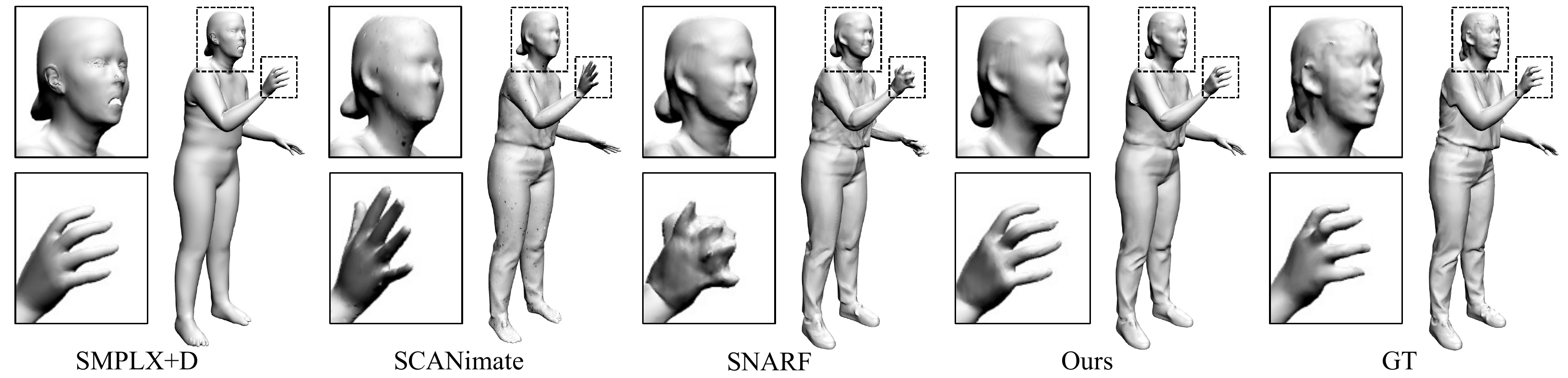}
\vspace{-3em}

\end{center}
   \caption{\textbf{Qualitative animation comparison on \ourdataset (Scans).} SMPLX+D fails to model face and garment details. SCANimate and SNARF generate poor hands (static or incomplete). Our method produces the most plausible face and hands, and keeps the clothing details comparable to strong baselines.}
\label{fig:cmp_real}
\end{figure*}
}

\newcommand{\figuredemoRealscan}{

\begin{figure}[h]
\begin{center}
\raggedleft
\includegraphics[width=\linewidth,trim=5 0 0 10,clip]{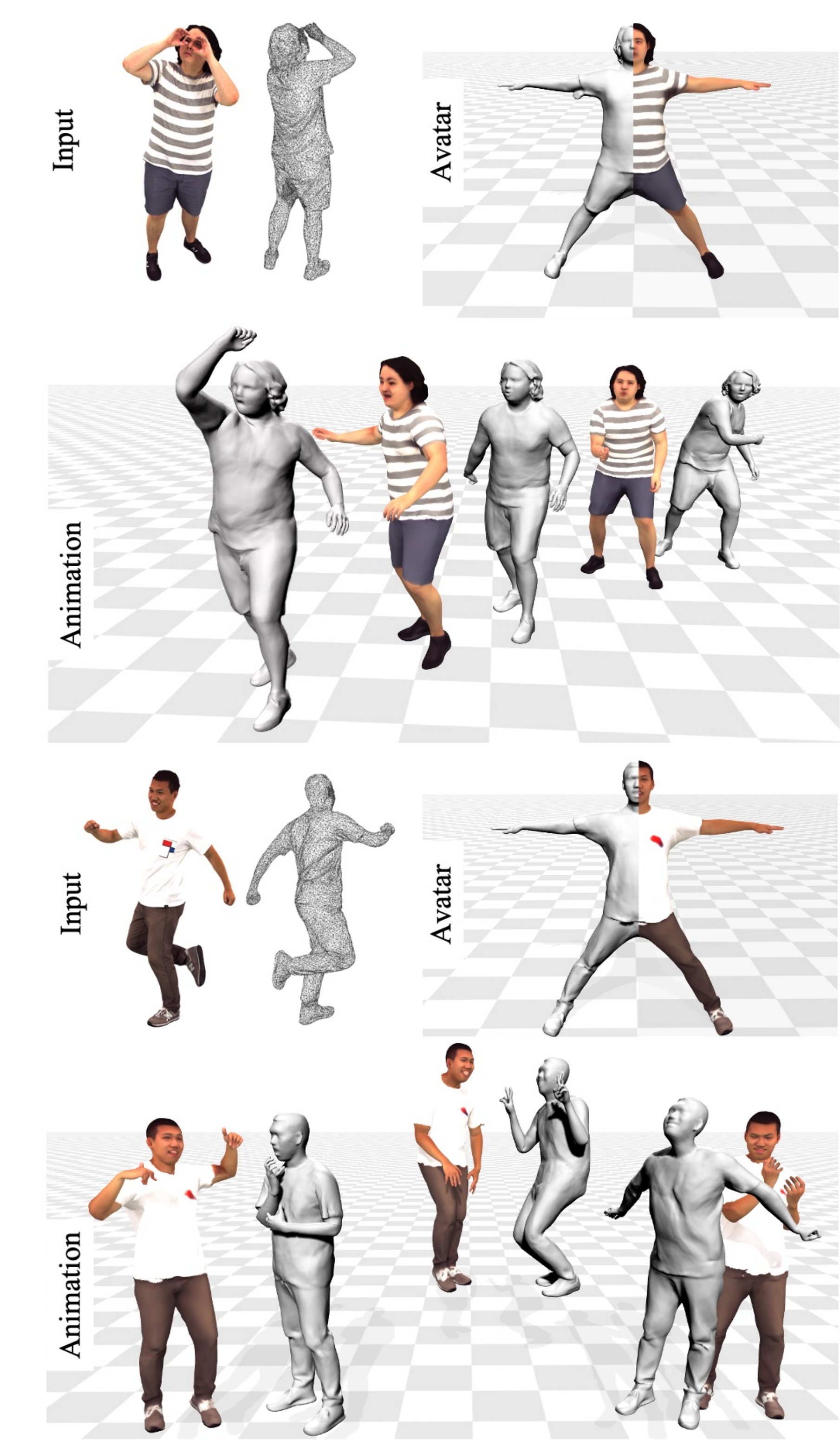}
\vspace{-3em}

\end{center}
   \caption{\textbf{Animation demonstration on \ourdataset (Scans).} Our method can handle relatively complex clothing patterns, hair styles, and varied facial expressions, hand, and body poses.}
\label{fig:demo_real}
\end{figure}
}

\newcommand{\figurecmpsyntheticRGBD}{

\begin{figure}[t]
\begin{center}
\raggedleft
\includegraphics[width=\linewidth,trim=0 0 0 0,clip]{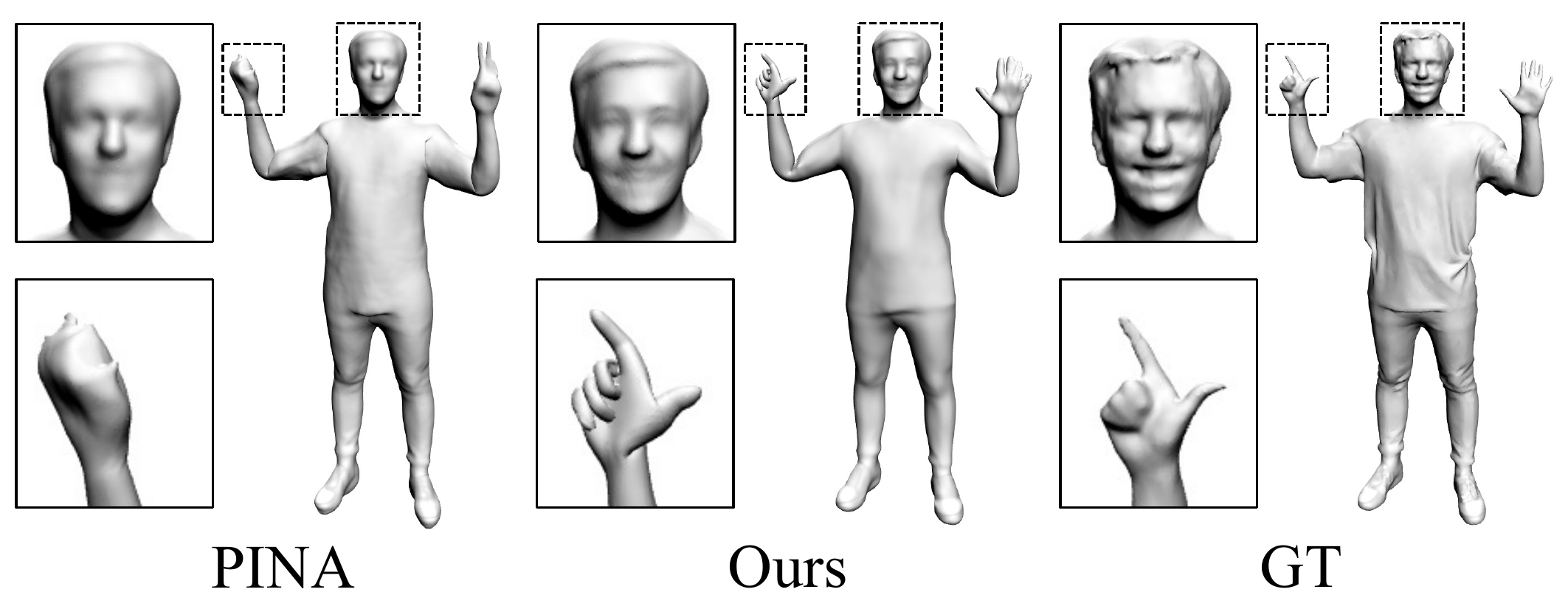}
\vspace{-3em}

\end{center}
   \caption{\textbf{\methodnames created from RGB-D input compared to PINA.} Notice how we obtain better hand and face geometry.}
\label{fig:cmp_RGBD}
\vspace{-0.5cm}
\end{figure}
}


\begin{abstract}
\vspace{-1em}
\setlength{\topsep}{0pt}
\setlength{\parskip}{.5ex}
\renewcommand{\floatsep}{1ex}
\renewcommand{\textfloatsep}{1ex}
\renewcommand{\dblfloatsep}{1ex}
\renewcommand{\dbltextfloatsep}{1ex}

We present \methodname, a novel avatar model that captures the full expressiveness of digital humans to bring about life-like experiences in telepresence, AR/VR and beyond.
Our method models bodies, hands, facial expressions and appearance in a holistic fashion and can be learned from either full 3D scans or RGB-D data. 
To achieve this, we propose a part-aware learned forward skinning module that can be driven by the parameter space of SMPL-X, allowing for expressive animation of \methodnames.
To efficiently learn the neural shape and deformation fields, we propose novel part-aware sampling and initialization strategies.
This leads to higher fidelity results, especially for smaller body parts while maintaining efficient training despite increased number of articulated bones.
To capture the appearance of the avatar with high-frequency details, we extend the geometry and deformation fields with a texture network that is conditioned on pose, facial expression, geometry and the normals of the deformed surface.
We show experimentally that our method outperforms strong baselines in both data domains both quantitatively and qualitatively on the animation task.
To facilitate future research on expressive avatars we contribute a new dataset, called \ourdataset, containing 233 sequences of high-quality textured scans from 20 participants, totalling 35,500 data frames. Project page: \href{https://skype-line.github.io/projects/X-Avatar/}{https://skype-line.github.io/projects/X-Avatar/}.
\end{abstract}

\vspace{-1.5em}
\section{Introduction}

A significant part of human communication is non-verbal in which body pose, appearance, facial expressions, and hand gestures play an important role. 
Hence, it is clear that the quest towards immersive, life-like remote telepresence and other experiences in AR/VR, will require methods to capture the richness of human expressiveness in its entirety.
Yet, it is not clear how to achieve this. Non-verbal communication involves an intricate interplay of several articulated body parts at different scales, which makes it difficult to capture and model algorithmically.

Parametric body models such as the SMPL family \cite{SMPL:2015, MANO:SIGGRAPHASIA:2017, SMPL-X:2019} have been instrumental in advancing the state-of-the-art in modelling of digital humans in computer vision and graphics. However, they rely on mesh-based representations and are limited to fixed topologies and in resolution of the 3D mesh. These models are focused on minimally clothed bodies and do not model garments or hair. Hence, it is difficult to capture the full appearance of humans.

Neural implicit representations hold the potential to overcome these limitations. 
Chen et al.~\cite{chen2021snarf} introduced a method to articulate human avatars that are represented by continuous implicit functions combined with learned forward skinning. This approach has been shown to generalize to arbitrary poses. While SNARF~\cite{chen2021snarf} only models the major body bones, other works have focused on creating implicit models of the face \cite{zheng2022IMavatar, Gafni2021Face}, the hands \cite{corona2022lisa}, or how to model humans that appear in garments \cite{dong2022pina} and how to additionally capture appearance~\cite{Saito2021Scanimate,tiwari21neuralgif}. Although neural implicit avatars hold great promise, to date no model exists that holistically captures the body and all the parts that are important for human expressiveness jointly. 

In this work, we introduce \methodname, an animatable, implicit human avatar model that captures the shape, appearance and deformations of complete humans and their hand poses, facial expressions, and clothing.
To this end we adopt the full-body pose space of SMPL-X \cite{SMPL-X:2019}. This causes two key challenges for learning \methodnames from data: (i) the significantly increased number of involved articulated body parts (9 used by \cite{chen2021snarf} vs. 45 when including hands and face) and (ii) the different scales at which they appear in observations. The hands and the face are much smaller in size compared to the torso, arms and legs, yet they exhibit similar or even more complex articulations.

\methodname consists of a shape network that models the geometry in canonical space and a deformation network to establish correspondences between canonical and deformed space via learned linear blend skinning (LBS). The parameters of the shape and deformation fields must be learned only from posed observations. SNARF~\cite{chen2021snarf} solves this via iterative correspondence search. This optimization problem is initialized by transforming a large number of candidate points via the bone transformations.
Directly adopting SNARF and initializing root-finding with only the body bones leads to poor results for the hands and face. Hence, to account for the articulation of these smaller body parts, their bone transformations must also be considered. 
However, correspondence search scales poorly with the number of bones, so na\"ively adding them makes training slow.
Therefore, we introduce a \textit{part-aware initialization} strategy which is almost 3 times faster than the na\"ive version while outperforming it quantitatively.
Furthermore, to counteract the imbalance in scale between the body, hands, and face, we propose a \textit{part-aware sampling} strategy, which increases the sampling rate for smaller body parts.
This significantly improves the fidelity of the final result.
To model the appearance of \methodnames, we extend the shape and deformation fields with an additional appearance network, conditioned on pose, facial expression, geometry and the normals in deformed space. All three neural fields are trained jointly.

\methodname can learn personalized avatars for multiple people and from multiple input modalities.
To demonstrate this, we perform several experiments. First, we compare our method to its most related work (SCANimate \cite{Saito2021Scanimate}, SNARF \cite{chen2021snarf}) on the GRAB dataset \cite{GRAB:2020,Brahmbhatt2019ContactDB} on the animation task of minimally clothed humans.
Second, we contribute a novel dataset consisting of 233 sequences of 20 clothed participants recorded in a high-quality volumetric capture stage \cite{MRCS}. The dataset consists of subjects that perform diverse body and hand poses (\eg, counting, pointing, dancing) and facial expressions (\eg, laughing, screaming, frowning).
On this dataset we show that \methodname can learn from 3D scans and (synthesized) RGB-D data. Our experiments show that \methodname outperforms strong baselines both in quantitative and qualitative measures in terms of animation quality.
In summary, we contribute:
\begin{compactitem}
    \item \methodname, the first expressive implicit human avatar model that captures body pose, hand pose, facial expressions and appearance in a holistic fashion.
    \item Part-aware initialization and sampling strategies, which together improve the quality of the results and keep training efficient.
    \item \ourdataset, a new dataset consisting of 233 sequences, of high-quality textured scans showing 20 participants with varied body and hand movements, and facial expressions, totalling 35,500 frames. 
\end{compactitem}
Data, models and SMPL[-X] registrations, will be released for scientific purposes.

\section{Related Work}
\paragraph{Explicit Human Models}
Explicit models use a triangulated 3D mesh to represent the underlying shape and are controlled by a lower-dimensional set of parameters.
Some models focus on capturing a specific part of the human, \eg, the body \cite{SMPL:2015, anguelov2005scape, osman2020star}, the hands \cite{MANO:SIGGRAPHASIA:2017}, or the face \cite{Blanz99BaselFaceModel,Li2017FLAME}, while others treat the human more holistically like we do in this work, \eg \cite{SMPL-X:2019, xu2020ghum,zanfir2020weakly,Osman2022SUPR,Joo2018TotalCapture}.
Explicit models are popular because the 3D mesh neatly fits into existing computer graphics pipelines and because the low-dimensional parameter space lends itself well for learning.
Only naturally have such models thus been applied to tasks such as RGB-based pose estimation \cite{PIXIE:2021, Sun2021ROMP,Kocabas2021PARE,Li2021hybrik,Choutas2020ExPose,song2020lgd, kolotouros2019SPIN,Yuan2022GLAMR,Sun2022BEV,Kanazawa2018HMR, pymafx2022, pymaf2021}, RGB-D fitting \cite{Tao2018DoubleFusion,Chen2016RealtimeRO,Bogo2015DetailedFR}, fitting to body-worn sensor data \cite{vonMarcard20183DPW,DIP:SIGGRAPHAsia:2018,Yi2022PIP}, or 3D hand pose estimation \cite{Hasson2019Learning,boukhayma20193d} with a resounding success.
Because the SMPL family does not natively model clothing, researchers have investigated ways to extend it, \eg via fixed additive 3D offsets \cite{Alldieck2018SMPLclothing, alldieck2019learning}, also dubbed SMPL+D, pose-dependent 3D offsets \cite{ma2020cape}, by modelling 3D garments and draping them over the SMPL mesh \cite{corona2021smplicit,bhatnagar2019mgn} or via local small surface patches \cite{Ma2021SCALE}.
Explicit models have seen a trend towards unification to model human expressiveness, \eg SMPL-X \cite{SMPL-X:2019} and Adam \cite{Joo2018TotalCapture}. \methodname shares this goal, but for implicit models.

\paragraph{Implicit Human Models}
Explicit body models are limited by their fixed mesh topology and resolution, and thus the expressive power required to model clothing and appearance necessitates extending these models beyond their original design. 
In contrast, using implicit functions to represent 3D geometry grants more flexibility.
With implicit models, the shape is defined by neural fields, typically parameterized by MLPs that predict signed distance fields \cite{park2019deepsdf}, density \cite{mildenhall2020nerf}, or occupancy \cite{mescheder2019occupancy} given a point in space.
To extend this idea to articulated shapes like the human body, NASA \cite{Deng2019NASA} used per body-part occupancy networks \cite{mescheder2019occupancy}. This per-part formulation creates artifacts, especially for unseen poses, which works such as \cite{chen2021snarf,Mihajlovic2021LEAP,Mihajlovi2022COAP} improve. SNARF \cite{chen2021snarf} does so via a forward warping field which is compatible with the SMPL \cite{SMPL:2015} skeleton, learns pose-independent skinning and generalizes well to unseen poses and people in clothing.
Other works \cite{Saito2021Scanimate,tiwari21neuralgif} model appearance and are learned from scans. \cite{xiu2022icon} creates avatars from video by relying on normals extracted from a 3D body model fitted to images and \cite{dong2022pina} does so from RGB-D video.

Moving beyond bodies, other work has investigated implicit models for faces \cite{zheng2022IMavatar,yenamandra2020i3dmm,ramon2021h3d,Gafni2021Face} and hands \cite{corona2022lisa}.
Yet, an implicit model that incorporates body, hands, face, and clothing in a single model is missing. \methodname fills this gap.
We do so by adopting neural forward skinning \cite{chen2021snarf} driven by SMPL-X \cite{SMPL-X:2019}. This seemingly simple change necessitates non-trivial improvements to the correspondence search as otherwise the iterative root finding is too slow and leads to poor results which we show empirically.
We propose to do so by introducing part-aware initialization and sampling strategies, which are incorporated into a single model.
Similar to \cite{Saito2021Scanimate}, we obtain color with an MLP that is fed with canonical points and conditioned on the predicted geometry. Thanks to the part-aware sampling strategy, our method produces higher quality results than \cite{Saito2021Scanimate} for the hands and faces.
Furthermore, in contrast to \cite{Saito2021Scanimate,tiwari21neuralgif,chen2021snarf}, \methodnames can be fit to 3D scans \emph{and} RGB-D videos.

\paragraph{Human Datasets}
Publicly available datasets that show the full range of human expressiveness and contain clothed and textured ground-truth are rare. GRAB \cite{GRAB:2020}, a subset of AMASS \cite{AMASS:ICCV:2019}, contains minimally clothed SMPL-X registrations. BUFF \cite{Zhang2017BUFF} and CAPE \cite{ma2020cape} do not model appearance and facial expressions. The CMU Panoptic Studio \cite{Joo2015Panoptic} dataset was used to fit Adam \cite{Joo2018TotalCapture} which does model hands and faces, but is neither textured nor clothed. Also, \cite{Joo2015Panoptic} does not contain scans. To study \methodnames on real clothed humans, we thus contribute our own dataset, \ourdataset which contains 35,500 frames of high-quality, texturized scans of real clothed humans with corresponding SMPL[-X] registrations.

\section{Method}
We introduce \methodname, a method for the modeling of implicit human avatars with full body control including body movements, hand gestures, and facial expressions. For an overview, please refer to \figref{fig:pipeline} and \figref{fig:deformer}.
Our model can be learned from two types of inputs, \ie, 3D posed scans and RGB-D images. We first recap the SMPL-X full body model. Then we describe the \methodname formulation, training scheme, and our part-aware initialization and sampling strategies.
For simplicity, we discuss the scan-based version without loss of generality and list the differences to depth-based acquisition in the \suppmat. 

\figurePipeline
\figureDeformer

\subsection{Recap: SMPL-X Unified Human Body Model}
Our goal is to create fully controllable human avatars. We use the parameter space of SMPL-X \cite{SMPL-X:2019}, which itself extends SMPL to include fully articulated hands and an expressive face. SMPL-X is defined by a function $M(\pose, \shape, \face): \mathbb{R}^{|\pose|} \times \mathbb{R}^{|\shape|} \times \mathbb{R}^{|\face|} \rightarrow \mathbb{R}^{3N}$, parameterized by the shape $\shape$, whole body pose $\pose$ and facial expressions $\face$. The pose can be further divided into the global pose $\pose_{g}$, head pose $\pose_{f}$, articulated hand poses $\pose_{h}$,
and remaining body poses $\pose_{b}$. Here, $|\pose_{g}|=3$, $|\pose_{b}|=63$, $|\pose_{h}|=90$, $|\pose_{f}|=9$, $|\shape|=10$, $|\face|=10$, $N=10,475$.

\subsection{Implicit Neural Avatar Representation}
\label{sec:avatar_model}
To deal with the varying topology of clothed humans and to achieve higher geometric resolution and increased fidelity of overall appearance, \methodname proposes a human model defined by articulated neural implicit surfaces. We define three neural fields: one to model the geometry via an implicit occupancy network, one to model deformation via learned forward linear blend skinning (LBS) with continuous skinning weights, and one to model appearance as an RGB color value.

\paragraph{Geometry}
We model the geometry of the human avatar in the canonical space with an MLP that predicts the occupancy value $f_{\text{occ}}$ for any 3D point $\mathbf{x}_c$ in this space. 
To capture local non-rigid deformations such as facial or garment wrinkles, we condition the geometry network on the body pose $\pose_{b}$ and facial expression coefficients $\face$.
We found empirically that high-frequency details are preserved better if positional encodings \cite{niemeyer2019occupancy} are applied to the input. Hence, the shape model $f_{\text{occ}}$ is denoted by:
\begin{equation}
\label{eq:1}
    f_{\text{occ}}: \mathbb{R}^{3} \times \mathbb{R}^{|\pose_{b}|} \times \mathbb{R}^{|\face|} \rightarrow [0,1] .
\end{equation}
The canonical shape is defined as the 0.5 level set of $f_{\text{occ}}$:
\begin{equation}
\label{eq:2}
     \mathcal{S} = \{ \mathbf{\ x}_c \ |\ f_{\text{occ}}(\mathbf{x}_c, \pose_{b}, \face) = 0.5 \ \}.
\end{equation}

\paragraph{Deformation}
To model skeletal deformation, we follow previous work \cite{chen2021snarf, li2022tava, dong2022pina, zheng2022IMavatar} and represent the skinning weight field in the canonical space by an MLP:

\begin{equation}
\label{eq:3}
f_{w}: \mathbb{R}^{3}  \rightarrow  \mathbb{R}^{n_{b}} \times \mathbb{R}^{n_{h}}\times \mathbb{R}^{n_{f}} ,
\end{equation}
where $n_{b}, n_{h}, n_{f}$ denotes the number of body, finger, and face bones respectively. Similar to \cite{chen2021snarf}, we assume a set of bones $\mathbf{G}$ and require the weights $\mathbf{w} \in \mathbb{R}^{|\mathbf{G}|}$ to fulfill $w_{i}\geq0$ and $\sum_{i}{w_i}=1$.
With the learned deformation field $\mathbf{w}$ and given bone transformations $\mathbf{B}=\{\mathbf{B}_1,...,\mathbf{B}_{|\mathbf{G}|}\}$, for each point $\mathbf{x}_c$ in the canonical space, its deformed counterpart is then uniquely determined:
\begin{equation}
\label{eq:4}
    \mathbf{x}_d =\mathbf{d}_w(\mathbf{x}_c, \mathbf{B}) = \sum_{i = 1}^{|\mathbf{G}|} w_i(\mathbf{x}_{c}) \mathbf{B}_i \, \mathbf{x}_c .
\end{equation}
Note that the canonical shape is a-priori unknown and learned during training. Since the relationship between deformed and canonical points is only implicitly defined, we follow \cite{chen2021snarf} and employ correspondence search.
We use Broyden's method~\cite{broyden1965class} to find canonical correspondences $\mathbf{x}_c$ for each deformed query point $\mathbf{x}_d$ iteratively as the roots of $\mathbf{d}_w(\mathbf{x}_c, \mathbf{B})-\mathbf{x}_d=0$.
In cases of self-contact, multiple valid solutions exist. Therefore the optimization is initialized multiple times by transforming deformed points $\mathbf{x}_d$ rigidly to the canonical space with each bone transformation. Finally, the set of valid correspondences $\mathcal{X}_c$ is determined via analysis of the local convergence.

\paragraph{Part-Aware Initialization}
At the core of our method lies the problem of jointly learning the non-linear deformations introduced by body poses \emph{and} dexterous hand articulation \emph{and} facial expressions. 
The above method to attain multiple correspondences scales poorly with the number of bones. 
Therefore, na\"{i}vely adding finger and face bones of SMPL-X to the initialization procedure, causes prohibitively slow training. Yet our ablations show that these are required for good animation quality (\cf. \tabref{tab:ablation study}).
Hence, we propose a part-aware initialization strategy, in which we first separate all SMPL-X bones $\mathbf{G}$ into four groups $\mathbf{G}_{B}$, $\mathbf{G}_{LH}$, $\mathbf{G}_{RH}$, $\mathbf{G}_{F}$. For a given deformed point with part label $\ell$, we then initialize the states $\{\mathbf{x}_{ci}^{0}\}$ and Jacobian matrices $\{\mathbf{J}_{ci}^{0}\}$ as:
\begin{equation}
\label{eq:5}
    \mathbf{x}_{ci}^{0} = \mathbf{B}_i^{-1} \cdot \mathbf{x}_{d}, \quad \mathbf{J}_{ci}^0=\left.\frac{\partial \mathbf{d}_{w}(\mathbf{x}, \mathbf{B}_i)}{\partial \mathbf{x}}\right|_{\mathbf{x}=\mathbf{x}_{ci}^0}, 
    i \in\mathbf{G}_{\ell} .
\end{equation}

We explain how we obtain the label $\ell$ for each point further below. The final occupancy prediction is determined via the maximum over all valid candidates $\mathcal{X}_{c} =\{\hat{\mathbf{x}}_{ci}\}_{i=1}^{|\mathbf{G}_{\ell}|}$:
\begin{equation}
\label{eq:6}
    o(\mathbf{x}_{d}, \pose_{b}, \face) = \max_{\hat{\mathbf{x}}_{c} \in \mathcal{X}_{c}} \{ f_{\text{occ}}(\hat{\mathbf{x}}_{c}, \pose_{b}, \face)\} .
\end{equation}
The correspondence in canonical space is given by:
\begin{equation}
    \mathbf{x}_c^{*} = \argmax_{\hat{\mathbf{x}}_{c} \in \mathcal{X}_{c}} \{ f_{\text{occ}}(\hat{\mathbf{x}}_{c}, \pose_{b}, \face)\} .
\end{equation}

This part-aware initialization is based on the observation that a point close to a certain body part is likely to be mostly affected by the bones in that part.
This scheme effectively creates four deformer networks, as shown in \figref{fig:pipeline}. However, note that all deformers share the same skinning weight network $f_w$ as highlighted in \figref{fig:deformer}. The only difference between them is how the iterative root finding is initialized.

\paragraph{Part-Aware Sampling}
Because hands and faces are comparatively small, while still exhibiting complex deformations, we found that a uniform sampling strategy for points $\mathbf{x}_d$ leads to poor results (cf. \tabref{tab:ablation study}, \figref{fig:ablation_hand}).
Hence, we further propose a part-aware sampling strategy, to over-sample points per area for small body parts.
Assuming part labels $\mathcal{P} = \{F, LH, RH, B \}$, for each point $\mathbf{p}_i$ in the 3D scan, we first find the closest SMPL-X vertex $\mathbf{v}_i$ and store its pre-computed body part label $k_i \in \mathcal{P}$.
Then, for each part $\ell \in \mathcal{P}$ we extract all points $\{ \mathbf{p}_i \mid k_i = \ell \}$ and re-sample the resulting mesh with a sampling rate specific to part $\ell$ to obtain $N_\ell$ many deformed points $\{\mathbf{x}_{di}\}_{i=1}^{N_\ell}$ for training.

\paragraph{LBS regularization}
To further account for the lower resolution and smaller scale of face and hands, we regularize the LBS weights of these parts to be close to the weights given by SMPL-X. A similar strategy has also been used by \cite{zheng2022IMavatar}. Our ablations show that this greatly increases the quality of the results (\cf \secref{sec:ablation_opt}).

\paragraph{Texture}
Similar to \cite{Saito2021Scanimate,tiwari21neuralgif} we introduce a third neural texture field to  predict  RGB values in canonical space.
Its output is the color value $c(\mathbf{x}_c, \mathbf{n}_d, \mathcal{F},  \pose_{b}, \face)$. 
This is, in addition to pose and facial expression, the color depends on the last layer $\mathcal{F}$ of the geometry network and the normals $\mathbf{n}_d$ in deformed space. 
This conditions the color prediction on the deformed geometry and local high-frequency details, which has been shown to be helpful \cite{Chen2022gDNA,zheng2022IMavatar}. Following \cite{zheng2022IMavatar}, the normals are obtained via $\mathbf{n}_d = \nabla_{\mathbf{x}_d} f_{occ}(\mathbf{x}^{*}_c, \pose_b, \face)$.
Therefore, the texture model $f_{\text{RGB}}$ is formulated as: 
\begin{equation}
\label{eq:7}
    f_{\text{RGB}}: \mathbb{R}^{3} \times \mathbb{R}^{3} \times 
    \mathbb{R}^{512}
    \times \mathbb{R}^{|\pose_{b}|} \times \mathbb{R}^{|\face|}
    \rightarrow \mathbb{R}^{3}.
\end{equation}

We apply positional encoding to all inputs to obtain better high-frequency details following best practices \cite{mildenhall2020nerf}.

\subsection{Training Process}
\label{sec:training}

\paragraph{Objective Function}
\label{sec:3.3}
For each 3D scan, we minimize the following objective:
\begin{equation}
\label{eq:8}
\begin{aligned}
\mathcal{L} = &
    \mathcal{L}_{occ} +
    \mathcal{L}_{RGB} +
    \mathcal{L}_{reg} .\\
\end{aligned}
\end{equation}
$\mathcal{L}_{occ}$ supervises the geometry and consists of two losses: the binary cross entropy loss $\mathcal{L}_{BCE}$ between the predicted occupancy $o(\mathbf{x}_{d}, \pose_{b}, \face)$ and the ground-truth value $o^{GT}(\mathbf{x}_d)$, and an L2 loss $\mathcal{L}_{n}$ on the normals:
\begin{equation}
\label{eq:9}
\begin{aligned}
\mathcal{L}_{occ} = &
\lambda_{BCE}\mathcal{L}_{BCE} +
\lambda_{n}\mathcal{L}_{n} \\
= & \lambda_{BCE} \sum_{\mathbf{x}_{d}\in \mathcal{P}_{\text{off}}} CE(o(\mathbf{x}_{d}, \pose_{b}, \face), o^{GT}(\mathbf{x}_d))\\
+&
\lambda_{n} \sum_{\mathbf{x}_{d}\in \mathcal{P}_{\text{on}}}
\begin{Vmatrix}
\mathbf{n}_d
-\mathbf{n}^{GT}(\mathbf{x}_d)
\end{Vmatrix}_{2} ,
\end{aligned}
\end{equation}
where $\mathcal{P}_{\text{on}}$, $\mathcal{P}_{\text{off}}$ separately denote points on the scan surface and points within a thin shell surrounding the surface \cite{dong2022pina}.
$\mathcal{L}_{RGB}$ supervises the point color:
\begin{equation}
\label{eq:10}
\mathcal{L}_{RGB} = \lambda_{RGB} \sum_{\mathbf{x}_{d}\in \mathcal{P}_{\text{on}}} 
\begin{Vmatrix}
c(\mathbf{x}_c, \mathbf{n}_d, \mathcal{F},  \pose_{b}, \face)
-c^{GT}(\mathbf{x}_d)
\end{Vmatrix}_{1} .
\end{equation}

Finally, $\mathcal{L}_{reg}$ represents the regularization term, consisting of the bone occupancy loss $\mathcal{L}_{\text{bone}}$, joint LBS weights loss $\mathcal{L}_{\text{joint}}$ and surface LBS weights loss $\mathcal{L}_{\text{surf}}$:
\begin{equation}
\label{eq:11}
\begin{aligned}
\mathcal{L}_{reg} = &
\lambda_{\text{bone}}\mathcal{L}_{\text{bone}} +
\lambda_{\text{joint}}\mathcal{L}_{\text{joint}} +
\lambda_{\text{surf}}\mathcal{L}_{\text{surf}}\\
= & \lambda_{\text{bone}} \sum_{\mathbf{x}_{c}\in \mathcal{P}_{\text{bone}}^{c}} CE(f_{\text{occ}}(\mathbf{x}_{c}, \pose_{b}, \face), 1)\\
+&
\lambda_{\text{joint}} \sum_{\mathbf{x}_{c}\in \mathcal{P}_{\text{joint}}^{c}}
\sum_{i \in \mathcal{N}(i)}
(w_{i}(\mathbf{x}_{c})
-0.5)^2
\\
+&
\lambda_{\text{surf}} \sum_{\mathbf{x}_{c}\in \mathcal{P}_{\text{surf}}^{c}}
\sum_{i \in \mathbf{G} \setminus \mathbf{G}_B}
(w_{i}(\mathbf{x}_{c})
-w_{i}^{GT}(\mathbf{x}_{c}))^2   , \\
\end{aligned}
\end{equation}
where $\mathcal{N}(i)$ are the neighboring bones of joint $i$ and $w_{i}^{GT}$ are the skinning weights taken from SMPL-X.
$\mathcal{L}_{reg}$ makes use of the supervision from registered SMPL-X meshes. For more details on the registration, please refer to the \suppmat. $\mathcal{P}_{\text{bone}}^{c}$, $\mathcal{P}_{\text{joint}}^{c}$, $\mathcal{P}_{\text{surf}}^{c}$ refer to points sampled on the SMPL-X bones, the SMPL-X joints and from the SMPL-X mesh surface respectively. The first two terms follow the definition of~\cite{chen2021snarf}. We add the last term to regularize the LBS weights for fingers and face which have low resolution and are more difficult to learn.

\newcommand{\tableablation}{
\begin{table*}[h]
    \centering
    \resizebox{1.00\linewidth}{!}{
    \begin{tabular}{clcc|cc|cc|cc}
    \hline \multirow{2}{*}{ID} & \multirow{2}{*}{Method} & \multicolumn{2}{c}{CD$\downarrow$}  &\multicolumn{2}{c}{CD-MAX $\downarrow$}  & \multicolumn{2}{c}{NC $\uparrow$} & \multicolumn{2}{c}{IoU $\uparrow$ }\\
    \cline{3-10} && All & Hands & All & Hands & All & Hands & All & Hands\\
    \hline
    A1 & Ours (init w body bones) & $5.42$ & $5.05$ & $57.54$ & $25.10$ & $0.940$ & $0.824$ & $0.964$  & $0.812$\\
    A2 & Ours (init w all bones) & $4.55$ & $4.35$ & $44.86$ & $20.71$ & $0.945$ & $0.845$ & $\mathbf{0.974}$ & $0.811$\\
    A3 & Ours (w/o part-aware sampling) & $4.68$ & $4.81$ & $47.51$ & $20.88$ & $0.947$ & $0.840$ & $0.972$  & $0.810$\\
    A4 & Ours (w/o LBS reg.) & $4.98$ & $7.27$ & $57.11$ & $43.38$ & $0.940$  & $0.797$ & $0.968$  & $0.768$ \\
    \hline
    A & Ours (complete) & $\mathbf{4.46}$ &
    $\mathbf{4.15}$ &
    $\mathbf{44.36}$ & 
    $\mathbf{20.61}$ &  
    $\mathbf{0.948}$ & $\mathbf{0.853}$ & $0.973$ & 
    $\mathbf{0.829}$ \\
    \hline
    \end{tabular}}
    \caption{\textbf{Ablation experiments for our major design choices.} We compute the metrics on the entire body (\textit{All}) and separately on the hands (\textit{Hands}) to better highlight the differences for the hands. All results are computed on a subset of \ourdataset (Scans). Our final model (A) only marginally outperforms A2, but is roughly 3 times faster to train. For qualitative comparisons please refer to \figref{fig:ablation_hand} and \figref{fig:ablation_part_aware_sampling}. }
    \label{tab:ablation study}
\end{table*}
}

\newcommand{\tableGRABAnimationfull}{
\begin{table}[h]
    \centering
    \resizebox{1.00\linewidth}{!}{
    \begin{tabular}{lcc|cc|cc|cc}
    \hline \multirow{2}{*}{Method} & \multicolumn{2}{c}{CD$\downarrow$}  &\multicolumn{2}{c}{CD-MAX $\downarrow$}  & \multicolumn{2}{c}{NC $\uparrow$} & \multicolumn{2}{c}{IoU $\uparrow$ }\\
    \cline{2-9} & All & Hands & All & Hands & All & Hands & All & Hands\\
    \hline SCANimate \cite{Saito2021Scanimate}& $2.60$ & $8.39$ & $54.75$ & $54.22$ & $0.967$ & $0.760$ & $0.941$ & $0.569$ \\
    SNARF \cite{chen2021snarf} & $1.37$ & $5.13$ & $33.86$ & $33.51$ & $0.977$ & $0.818$ & $0.967$ & $0.739$ \\
    \textbf{Ours} & $\mathbf{0.94}$ & $\mathbf{0.79}$ & $\mathbf{21.43}$ & $\mathbf{4.79}$ & $\mathbf{0.985}$ & $\mathbf{0.957}$ & $\mathbf{0.991}$ & $\mathbf{0.895}$ \\
    \hline
    \end{tabular}}
    \caption{\textbf{Quantitative results on GRAB dataset.}  Our method outperforms all baselines, especially for the hand part (\cf  \figref{fig:cmp_GRAB}).}
    \label{tab:animation results on GRAB}
    \vspace{-0.5cm}
\end{table}
}

\newcommand{\tableRealScanAnimation}{
\begin{table}[h]
    \centering
    \resizebox{1.00\linewidth}{!}{
    \begin{tabular}{lcc|cc|cc|cc}
    \hline \multirow{2}{*}{Method} & \multicolumn{2}{c}{CD$\downarrow$}  &\multicolumn{2}{c}{CD-MAX $\downarrow$}  & \multicolumn{2}{c}{NC $\uparrow$} & \multicolumn{2}{c}{IoU $\uparrow$ }\\
    \cline{2-9} & All & Hands & All & Hands & All & Hands & All & Hands\\
    \hline SMPLX+D & $5.75$ & $5.19$ & $48.41$ & $23.48$ & $0.921$ & $0.790$ & $0.957$ & $0.774$\\
    SCANimate \cite{Saito2021Scanimate} & $6.54$ & $9.78$ & $59.71$ & $48.32$ & $0.925$ & $0.726$ & $0.919$ & $0.557$\\
    SNARF \cite{chen2021snarf} & $5.05$ & $7.23$ & $55.06$ & $37.15$ & $0.934$ & $0.788$ & $0.937$ & $0.608$  \\
    \textbf{Ours} & $\mathbf{4.43}$ & $\mathbf{5.14}$ & $\mathbf{47.56}$ & $\mathbf{22.15}$ & $\mathbf{0.939}$ & $\mathbf{0.793}$ & $\mathbf{0.965}$ & $\mathbf{0.776}$  \\
    \hline
    \end{tabular}}
    \caption{\textbf{Quantitative results on \ourdataset (Scans).} We beat all baselines both for the entire body (\textit{All}) and hands only (\textit{Hands}).  }
    \label{tab:animation results on Real Dataset}
    \vspace{-0.5cm}
\end{table}
}

\newcommand{\tableRGBDAnimation}{
\begin{table}[h]
    \centering
    \resizebox{1.00\linewidth}{!}{
    \begin{tabular}{lcc|cc|cc|cc}
    \hline \multirow{2}{*}{Method} & \multicolumn{2}{c}{CD$\downarrow$}  &\multicolumn{2}{c}{CD-MAX $\downarrow$}  & \multicolumn{2}{c}{NC $\uparrow$} & \multicolumn{2}{c}{IoU $\uparrow$ }\\
    \cline{2-9} & All & Hands & All & Hands & All & Hands & All & Hands\\
    \hline PINA \cite{dong2022pina}& $5.41$ & $9.51$ & $66.05$ & $48.07$ & $0.928$ & $0.771$ & $0.910$ & $0.566$\\
    \textbf{Ours} &  $\mathbf{5.33}$ & $\mathbf{5.27}$ & $\mathbf{51.73}$ & $\mathbf{22.86}$ & $\mathbf{0.936}$ & $\mathbf{0.797}$ & $\mathbf{0.947}$ & $\mathbf{0.768}$   \\
    \hline
    \end{tabular}}
    \caption{\textbf{Quantitative results on \ourdataset (RGB-D).} Our method outperforms PINA in all metrics. Improvements are more pronounced for hands (\cf \figref{fig:cmp_RGBD} for visual comparison).
    }
    \label{tab:animation results on synthetic RGB-D video dataset}
\end{table}
}

\section{Experiments}
We first introduce the datasets and metrics that we use for our experiments in \secref{sec:datasets}. \secref{sec:ablation_opt} ablates all important design choices. In \secref{sec:baselines} we briefly describe the state-of-the-art methods to which we compare our method. Finally we show and discuss the results in \secref{sec:experiments_GRAB}-\ref{sec:experiments_RGBD}. We focus on the challenging animation task, hence all the comparisons are conducted on entirely unseen poses. For completeness, we also report reconstruction results in the \suppmat.

\tableablation
\figureablationhandgeometry

\subsection{Datasets}
\label{sec:datasets}

\paragraph{GRAB \cite{GRAB:2020}}
We use the GRAB subset of AMASS \cite{AMASS:ICCV:2019} for training and evaluate our model on SMPL-X meshes of minimally clothed humans. GRAB contains a diverse set of hand poses and facial expressions with several subjects. We pick the subject with the most pose variation and randomly select 9 sequences for training and three for validation. This results in 9,756 frames for training and 1,272 test frames.
\paragraph{\ourdataset (Scans)}
Currently, there exists no publicly available dataset containing textured 3D clothed scans of humans with a large variation of body poses, hand gestures and facial expressions. Therefore, we captured our own dataset, for which we leveraged a high-quality, multi-view volumetric capture stage \cite{MRCS}. We call the resulting dataset \ourdataset. It consists of 20 subjects (11 males, 9 females) with various clothing types and hair style. The collection of this dataset has been approved by an internal ethics committee. For each subject, we split the motion sequences into a training and test set. In total, there are 29,036 poses for training and 6,439 test poses. \ourdataset also contains ground-truth SMPL-X parameters, obtained via a custom SMPL-X registration pipeline specifically designed to deal with low-resolution body parts. More details on the registration process and contents of \ourdataset are in \suppmat.
\paragraph{\ourdataset (RGB-D)}
We take the textured and posed scans from \ourdataset and render them to obtain corresponding synthetic RGB-D images. For every time step, we render exactly one RGB-D image from a virtual camera, while the camera gradually rotates around the participant during the duration of the sequence. This is, the RGB-D version of \ourdataset contains the same amount of frames as the scan version in both the training and test set.

\paragraph{Metrics} We evaluate the geometric accuracy via volumetric IoU, Chamfer distance (CD) (mm) and normal consistency (NC) metrics, following the practice in PINA \cite{dong2022pina}. Because these metrics are dominated by large surface areas, we always report the metrics for the entire body (\textit{All}) and the hands separately (\textit{Hands}).

\subsection{Ablation Study}
\label{sec:ablation_opt}

\paragraph{Part-Aware Initialization}
The part-aware initialization for correspondence search is critical to accelerate training and to find good correspondences in small body parts. To verify this, we compare with two variations adapted from SNARF \cite{chen2021snarf}. First, (A1) initiates the optimization states only via the body's bone transformations, while (A2) initializes using all bones (body, hands, face).
\textbf{Results:} A1 suffers from strong artifacts for hands and the jaw (\cf \figref{fig:ablation_hand} and \tabref{tab:ablation study}, A1).
The final model is 3 times faster than A2 ($0.7$ iterations per second vs. $0.25$), yet it still retains high fidelity and even outperforms A2 by a small margin (\cf \tabref{tab:ablation study}, A2 and the \suppmat. for qualitative results). Thus we conclude that part-based initialization of the deformer is an efficient way to find accurate correspondences.

\paragraph{Part-Aware Sampling}
To verify the importance of part-aware sampling, we compare our model to a uniform sampling baseline (A3). \textbf{Results:} This component has two effects: a) it strongly improves the hand shape (\cf second column of \figref{fig:ablation_hand} and \tabref{tab:ablation study}, A3) and b) it improves texture details in the eye and mouth region (\cf \figref{fig:ablation_part_aware_sampling}).

\figureablationpartawaresampling

\paragraph{LBS Weights Regularization for Hands and Face}
The first column in \figref{fig:ablation_hand} shows that without regularizing the learned LBS weights with the SMPL-X weights, the learned hand shape is poor. This is further substantiated by a 75\% increase in Chamfer distance for the hand region, compared to our final method (\cf \tabref{tab:ablation study}, A4).

\subsection{Baselines}
\label{sec:baselines}

\paragraph{Scan-based methods}
We compare our 3D scan-based method variant on both GRAB and \ourdataset to SMPLX+D, SCANimate and SNARF  baselines. We adapt SMPLX+D from SMPL+D introduced in \cite{bhatnagar2020ipnet}. This baseline uses an explicit body model, SMPL-X, and models clothing with additive vertex offsets. To compare with SCANimate and SNARF, we use publicly available code. For details on the baselines, we refer to the \suppmat.

\paragraph{RGB-D Video-based methods}
We compare our RGB-D method variant on the \ourdataset (RGB-D) dataset to PINA \cite{dong2022pina}, a SMPL-based implicit human avatar method learned from RGB-D inputs. We assume that the ground truth pose and shape are known. For a fair comparison we do not optimize these parameters in PINA.

\subsection{Results on GRAB Dataset}
\label{sec:experiments_GRAB} 
\tabref{tab:animation results on GRAB} summarizes results on the GRAB dataset. Overall, our method beats all baselines, especially for the hands, where the margin is large. \figref{fig:cmp_GRAB} visually shows that the quality of the hands and face learned by our method is much higher: SCANimate learns a mean hand and SNARF generalizes badly to the unseen poses. Since GRAB meshes are minimally clothed, we omit SMPLX+D from  comparison.

\figurecmpGRAB
\tableGRABAnimationfull

\subsection{Results on \ourdataset (Scans)}

\tabref{tab:animation results on Real Dataset} shows that our method also outperforms the baselines on \ourdataset. \figref{fig:cmp_real} qualitatively shows differences. SMPLX+D, limited by its fixed topology and low resolution, cannot model details like hair and wrinkles in clothing. SCANimate and SNARF are SMPL-driven, so they either learn a static or incomplete hand. Our method balances the different body parts so that hands are well-structured, but also the details on the face and body are maintained. \figref{fig:teaser} and \figref{fig:demo_real} show more animation results. 

\figurecmpRealscan
\tableRealScanAnimation

\tableRGBDAnimation
\figurecmpsyntheticRGBD

\subsection{Results on \ourdataset (RGB-D)}
\label{sec:experiments_RGBD}

\tabref{tab:animation results on synthetic RGB-D video dataset} shows our model's performance compared to PINA \cite{dong2022pina}. Our method outperforms PINA on all metrics. \figref{fig:cmp_RGBD} further qualitatively shows that without utilizing the hand and face information in the modelling process, the face and hands produced by PINA are not consistent with the input pose. Our model generates a) more realistic faces as the shape network is conditioned on facial expression and b) better hand poses because we initialize the root finding with hand bone transformations. 

\figuredemoRealscan

\section{Conclusion}

\textbf{Limitations} \quad \methodname struggles to model loose clothing that is far away from the body (\eg skirts).
Furthermore, generalization capability beyond a single person is still limited, \ie we train one model for each subject.

\textbf{Conclusion} \quad
We propose \methodname, the first expressive implicit human avatar model that captures body pose, hand pose, facial expressions and appearance in a holistic fashion. We have demonstrated our method's expressive power, the benefit of our proposed part-aware initialization and sampling strategy, and the capability of creating it from multiple input modalities with the aid of our newly introduced \ourdataset dataset. We believe that our method along with \ourdataset will promote further scientific research in creating expressive digital avatars.

{\small
\bibliographystyle{ieee_fullname}
\bibliography{egbib}
}

\end{document}